\documentclass[10pt,twocolumn,letterpaper]{article}

\usepackage{iccv}
\usepackage{times}
\usepackage{epsfig}
\usepackage{graphicx}
\usepackage{amsmath}
\usepackage{amssymb}
\usepackage{makecell}
\usepackage{comment}
\usepackage{booktabs}
\usepackage{multirow}
\usepackage{color, colortbl}
\usepackage{floatrow}

\definecolor{Gray}{gray}{0.9}

\usepackage[pagebackref=true,breaklinks=true,letterpaper=true,colorlinks,bookmarks=false]{hyperref}

\iccvfinalcopy % *** Uncomment this line for the final submission

\def\iccvPaperID{33} % *** Enter the ICCV Paper ID here

\ificcvfinal\fi

\begin{document}

%%%%%%%%% TITLE
\title{4-Connected Shift Residual Networks}

\author{Andrew Brown, Pascal Mettes, and Marcel Worring\\
University of Amsterdam\\
{\tt\small a.g.brown,p.s.m.mettes,m.worring@uva.nl}
}

\maketitle
\ificcvfinal\thispagestyle{empty}\fi

\begin{abstract}

 The shift operation was recently introduced as an alternative to spatial convolutions. The operation moves subsets of activations horizontally and/or vertically. Spatial convolutions are then replaced with shift operations followed by point-wise convolutions, significantly reducing computational costs. In this work, we investigate how shifts should best be applied to high accuracy CNNs. We apply shifts of two different neighbourhood groups to ResNet on ImageNet: the originally introduced 8-connected (8C) neighbourhood shift and the less well studied 4-connected (4C) neighbourhood shift. We find that when replacing ResNet's spatial convolutions with shifts, both shift neighbourhoods give equal ImageNet accuracy, showing the sufficiency of small neighbourhoods for large images. Interestingly, when incorporating shifts to all point-wise convolutions in residual networks, 4-connected shifts outperform 8-connected shifts. Such a 4-connected shift setup gives the same accuracy as full residual networks while reducing the number of parameters and FLOPs by over 40\%. We then highlight that without spatial convolutions, ResNet's downsampling/upsampling bottleneck channel structure is no longer needed. We show a new, 4C shift-based residual network, much shorter than the original ResNet yet with a higher accuracy for the same computational cost. This network is the highest accuracy shift-based network yet shown, demonstrating the potential of shifting in deep neural networks.

\end{abstract}

%------------------------------------------------------------------------- 
\section{Introduction}
\label{sec:intro}

In recent years, convolutional neural networks (CNNs) have radically improved the state-of-the-art in image classification accuracy. Yet this improvement has come at an exponentially increasing computational expense. The popular CNN ResNet \cite{ResNet} (2016) is approximately ten times the computational cost of earlier ImageNet challenge \cite{ImageNet} winners such as AlexNet \cite{AlexNet} (2012). The most accurate networks on ImageNet today, SENet154 \cite{SqueezeExcite} (2018) and NasNet \cite{NasNet} (2018), are approximately twice as expensive as ResNet. 

This work focuses on optimising the computational footprint of high accuracy networks by replacing one of their costliest components in terms of parameters and FLOPs, the spatial convolution. We work with ResNet, as this popular network is still close to the most accurate networks today, and modify its architecture with the shift operation. The shift operation was recently introduced by Wu \textit{et al.} \cite{Zero} and moves all elements within a given channel's image plane horizontally and/or vertically, with different (groups of) channels undertaking different moves. The operation is FLOP and parameter free, being theoretically equivalent to a re-referencing of the initial activations maps \cite{Primitives}. Spatial convolutions are replaced by a shift followed by a point-wise ($1\times1$) convolution, itself equivalent to a simple matrix multiplication, and so require fewer parameters and FLOPs. 

\begin{figure*}
\centering
  \includegraphics[width=0.85\linewidth]{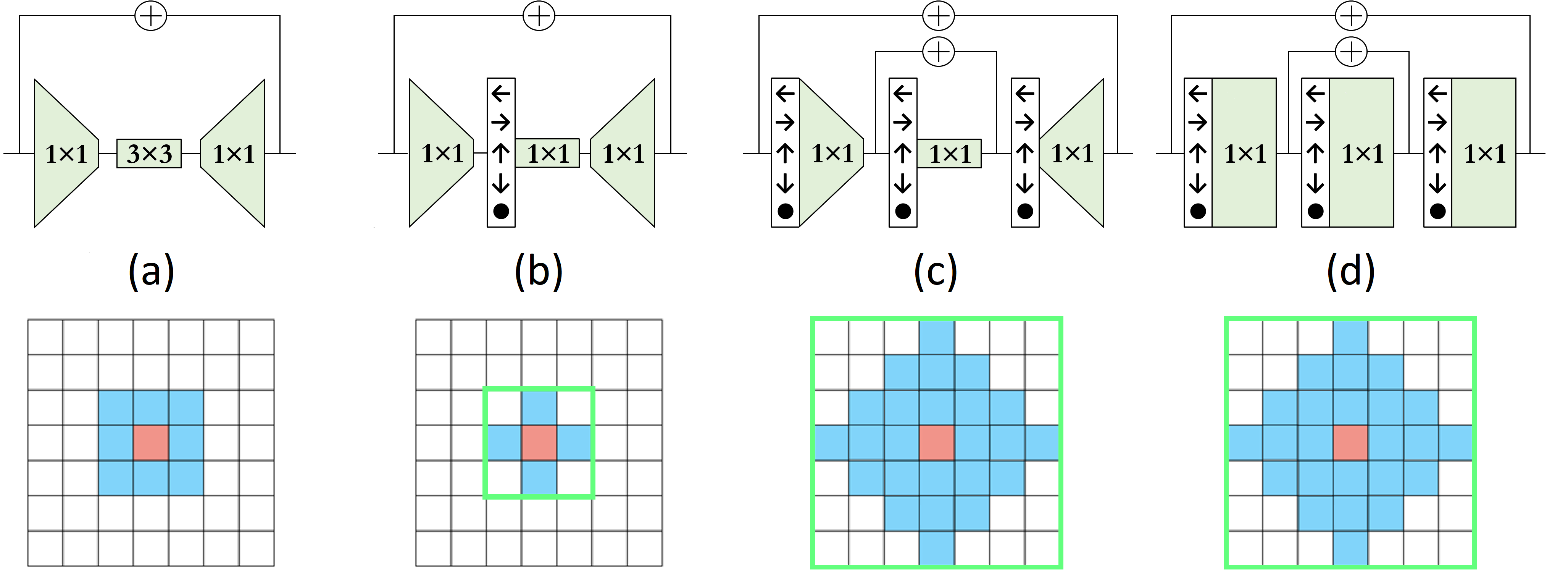}
  \caption{(Top) Original and proposed residual block designs. Green blocks are convolutions; the height of the blocks correspond to the number of channels. White boxes with arrows indicate shifts. Batch normalization and ReLU operations, applied after each convolution, are omitted for clarity. From left to right: (a) the down- and up- sampling bottleneck design of the original ResNet, (b) a single 4-connected shift residual network, with the bottleneck channel design (c) our proposed multi-shift block, with shifts applied before every $1 \times 1$ convolution, also with the bottleneck design, and (d) our simplified channel-flattened multi-shift residual block, which is enabled through the replacement of spatial convolutions. (Bottom) The theoretical receptive field extent of each residual block. Elements within the receptive field of each block are shown in blue, with the origin indicated in red. Green boxes indicate receptive field extents if 8-connected shifts are used instead of 4-connected shifts.}
  \label{fig:Fig1}
\end{figure*}

So far, several shift-based CNN architectures have been proposed \cite{Primitives,Active,Zero} for the small-scale image datasets CIFAR10 and CIFAR100 \cite{CIFAR}. Computationally constrained CNN architectures \cite{Primitives,Active,Zero} have also been proposed for large images on ImageNet \cite{ImageNet}. High accuracy shift network architectures, equalling or surpassing ResNet \cite{ResNet} on ImageNet, have not yet been explored. Here, we ask: for high accuracy networks such as ResNet, how should shifts be applied, and what discrete shift neighbourhood is best? The question of the spatial extent of neighbourhoods in visual recognition is a long-standing challenge, dating back to cellular arrays in image processing \cite{Cellular} and subsequently in image filtering \cite{Serra}. For the spatial extent in rectangular arrays, two neighbourhoods are generally employed: the 8-connected (8C) neighbours (left, right, up and down + diagonals); and the 4-connected (4C) neighbours (left, right, up and down only). These neighbourhoods, illustrated in Fig. \ref{fig:Fig1}, are also known as the \emph{Moore neighbourhood} and \emph{von Neumann neighbourhood}, respectively \cite{CVHandbook}. Here we look at which neighbourhood to use in the high-accuracy deep learning setting.

The main focus of this work is two-fold. First, we aim to employ shifts on a full ResNet network (ResNet101) on a large-scale image dataset. This is with a view to optimising network architecture, either by maintaining accuracy while cutting computational cost, or by maintaining computational cost while improving accuracy. Second, we investigate which neighbourhood extent is sufficient for image recognition in such networks. The original shift replaces the $3 \times 3$ spatial convolutional kernel with $3 \times 3$ shifts. Implicitly, they opt for the Moore neighbourhood, but is such a full extent necessary? 

In line with these focus points, we propose two extensions of the ResNet architecture. The first network adds multiple shifts to ResNet's residual blocks to reduce FLOPs and maintain accuracy. The second network focuses on accuracy for the same FLOPs. We highlight that, without the spatial convolutions, the 'bottleneck' in ResNet's channel structure is no longer needed. We construct a shorter network with a simpler channel structure, without down- and up-sampling, and show it gives superior performance on ImageNet. We then make the following contributions:

\begin{itemize}
  \item We explore alternative neighbourhoods variants of the shift operation in ResNet on ImageNet. When directly replacing spatial convolutions, we find that shifting only to the 4-connected neighbours is sufficient for image recognition. 
  \item We propose a multi-shift architecture, adding spatial information to the downsampling and up-sampling convolutions of ResNet. We find that performance is improved with this approach, but only for 4-connected shifts. This result highlights the importance of constraining neighbourhood extents when shifting on large networks. Our proposed multi-shift network then reduces ResNet's computational costs by 43\% while maintaining accuracy. 
  \item We propose a multi-shift-based ResNet variant without the 'bottleneck', which becomes possible when replacing spatial convolutions with shifts. The channel structure then becomes less complex, as the same number of channels is used throughout each residual block, and the network much shorter (35 layers) than the original (101 layers). We show a network with this design using 4-connected shifts which has approximately the same computational costs as ResNet101 and an accuracy increase of +0.8\%. This is the highest accuracy shift-based network ever demonstrated on ImageNet.
\end{itemize}

\section{Related Work}
\label{sec:related}

The shift operation was first introduced in \cite{Zero}. Shifts translate activation maps horizontally and/or vertically to a neighbouring position. The shift operations of \cite{Zero} consider a square 8-connected neighbourhood for image classification, with the shift-based CNN architectures demonstrated in \cite{Zero} primarily optimised for the miniature image datasets CIFAR10/100 \cite{CIFAR}. Computationally constrained networks for larger images, tested on ImageNet \cite{ImageNet}, were also shown. A higher accuracy network (ShiftResNet50) for ImageNet was also tested, but its architecture is unpublished. We estimate the FLOPs of this architecture and compare its results to ours in this work. We build on \cite{Zero}; we focus on varying the shift neighbourhood and applying it to high accuracy networks for large images.
We furthermore propose new residual architectures for shifts, resulting in competitive networks with low computational cost, outperforming other shift approaches, such as Wu et al. \cite{Zero}.

Other works have also investigated varying shifts operations for image classification. Closely related to this work is that of \cite{Primitives}, who vary the (discrete) neighbourhood of shifts for miniature images. They then build a compact model for large images (accuracy 67.0\%), though the FLOPs and parameters of this model were not shown. Comparatively, we focus on comparing shift neighbourhoods for the large image setting and for much larger, high accuracy networks (78.4\%). We explore additional shift variants, discussing when they are appropriate, showing the architectural changes required to optimise large shift networks.

Active-shifts, introduced in \cite{Active}, also relate to this work. By realising shifts as bi-linear interpolations of an input activation map, the horizontal and vertical motions of a shift can treated as real, trainable values.  However, active-shifts require additional FLOPs to calculate these interpolations. Further, as activation map motion is non-integer, active shifts always require additional activation map copies in any implementation \cite{Primitives}. \cite{Active} also focus on optimising network architectures for miniature image datasets and for computationally constrained models on ImageNet, while we go beyond small datasets and compact networks for shifting. 

Most recently, sparse-shifts were introduced in \cite{Sparse}. Sparse shifts attempt to learn discrete shift neighbourhoods by integer approximations of active-shifts. While \cite{Sparse} do consider constraining shift neighbourhoods through an L1 regularisation of shift magnitude, they ultimately find unconstrained shift-neighbourhoods to be optimal, in contrast to our results. We show the results of \cite{Sparse} in Fig. \ref{fig:Fig3}.

Shifts have also been applied in other contexts. These range from optimising shift-based CNNs for use with FPGAs \cite{Synetgy} or systolic arrays \cite{Systolic} to re-purposing shifts for new tasks such as video recognition \cite{MotionFeatureNetwork}. To our knowledge, no work has yet explored how shifts should be applied to a high accuracy network for image classification, or explored shift neighbourhoods in this setting. 

In a broader sense, our work relates to methods to reduce network computational cost of larger CNNs. Examples of such methods are in network design (e.g. \cite{MobileNetV2,Squeeze}), tensor decomposition (e.g. \cite{TensorFactorisation,TensorFactorisation2}), network pruning (e.g. \cite{NetworkPruning,NetworkPruning2}) and student-teacher network training (e.g. \cite{StudentTeacher,StudentTeacher2}). Shift operations in general and the shift-variants and CNN architectures we consider here are both complementary to and distinct from these approaches.

%\vspace{-8pt}

\section{Method}

%\vspace{-2pt}

\subsection{The 4-connected shift operation}

The shift operation, introduced in \cite{Zero}, moves all the elements of an activation map an integer number of elements along spatial directions. The set of allowed shift directions is the \textit{shift neighbourhood}. Different (subsets of) channels are moved to different positions within this neighbourhood. In the original design, the shift neighbourhood matches the square neighbourhood of the equivalent square convolutional kernel of a spatial convolution (Fig. \ref{fig:Fig1} a). Thus, to match a square kernel of spatial extent $D_k \times D_k$, there are $K=D_k^2$ neighbourhood positions. To perform the operation, an activation map of $M$ channels is split into $D_k^2$ subgroups, each of a size $M // D_k^2$ channels, where '$//$' denotes integer division. Each of the $K$ channel subgroups' activation maps is then moved in one of the $K$ neighbourhood positions. In the case that division $M // D_k^2$ is not exact, the remaining $(M\ \text{mod}\ D_k^2)$ channels are added to the origin (central element) subgroup and are, in practice, unmoved.

As stated in the introduction, here we ask: what shift neighbourhood is optimal? We compare the two neighbourhoods which have a long history of importance in image processing: the 8-connected (8C) Moore neighbourhood and the 4-connected (4C) von Neumann neighbourhood \cite{CVHandbook,8vs4_1,8vs4_2,8vs4_3}. Fig. \ref{fig:Fig1} visually compares these neighbourhoods. Most modern CNN frameworks, such as TensorFlow \cite{TensorFlow} or Pytorch \cite{Pytorch} only allow rectangular convolutional kernels in spatial convolution operations, and do not allow, for example, the cross shape of 4C neighbourhoods. Shift operations provide a new opportunity to study image neighbourhood connectivity, as they do not rely on these spatial convolution operations. 

More formally, for two point-sets $\mathbf{X}$ and $\mathbf{Z}$, corresponding to the input and output activation maps of a shift, we define the neighbourhood function $N$ from $\mathbf{X}$ to $\mathbf{Z}$ \cite{CVHandbook}:

%\vspace{-8pt}

\begin{equation}
N : \mathbf{X} \rightarrow 2^{\mathbf{Z}},
\end{equation}

%\vspace{-5pt}

such that for each point $\mathbf{x} \in \mathbf{X}$, it holds that $N(\mathbf{x}) \subset \mathbf{Z}$. The 8C neighbourhood and 4C neighbourhood functions are defined as:

%\vspace{-16pt}

\begin{equation}
\begin{split}
    N_{8C}(\mathbf{x}) = &\{\mathbf{y} : \mathbf{y} = (x_{1} \pm a,x_{2} \pm b) \ ,\\
    & a, b \in \{0,1\}\},
\end{split}
\end{equation}
\begin{equation}
\begin{split}
    N_{4C}(\mathbf{x}) = & \{\mathbf{y} : \mathbf{y} = (x_{1} \pm a,x_{2}) \ or \ \mathbf{y} = (x_{1},x_{2}\pm b),\\
    & a, b \in \{0,1\}\}.
\end{split}
\end{equation}

\begin{table*}[t]
\centering
\resizebox{\textwidth}{!}{
\begin{tabular}{|c|c|c|c|c|}
\hline
 Input shape & ResNet101 & Shift & Multi-shift & Flattened multi-shift\\
\hline
$224 \times 224$ & \multicolumn{4}{c|}{conv $7 \times 7$ stride-2, max-pool $3 \times 3$ stride-2, out: 64}\\
\hline
$56 \times 56$ & $\begin{bmatrix} \text{conv} 1 \times 1, \ 64 \\ \text{conv} 3 \times 3, \ 64 \\ \text{conv} 1 \times 1, 256 \\ \end{bmatrix}$ $\times$ \ 3 & $\begin{bmatrix} \phantom{\textbf{S}} \ \text{conv} 1 \times 1, \ 64 \\ \textbf{S} \ \text{conv} 1 \times 1, \ 64 \\ \phantom{\textbf{S}} \ \text{conv} 1 \times 1, 256 \\ \end{bmatrix}$ $\times$  \ 3 & $\begin{bmatrix} \textbf{S} \ \text{conv} 1 \times 1, \ 64 \\ \textbf{S} \  \text{conv} 1 \times 1, \ 64 \\ \textbf{S} \ \text{conv} 1 \times 1, 256 \\ \end{bmatrix}$ $\times$ \ 3 & $\begin{bmatrix} \textbf{S} \  \text{conv} 1 \times 1, 256 \\ \textbf{S} \  \text{conv} 1 \times 1, 256 \\ \textbf{S} \  \text{conv} 1 \times 1, 256 \\ \end{bmatrix}$ $\times$ 1\\
\hline
$56 \times 56$ & $\begin{bmatrix} \text{conv} 1 \times 1, 128 \\ \text{conv} 3 \times 3, 128 \\ \text{conv} 1 \times 1, 512 \\ \end{bmatrix}$ $\times$ \ 4 & $\begin{bmatrix} \phantom{\textbf{S}} \ \text{conv} 1 \times 1, 128 \\ \textbf{S} \ \text{conv} 1 \times 1, 128 \\ \phantom{\textbf{S}} \ \text{conv} 1 \times 1, 512 \\ \end{bmatrix}$ $\times$ \ 4 & $\begin{bmatrix} \textbf{S} \ \text{conv} 1 \times 1, 128 \\ \textbf{S} \ \text{conv} 1 \times 1, 128 \\ \textbf{S} \ \text{conv} 1 \times 1, 512 \\ \end{bmatrix}$ $\times$ \ 4 & $\begin{bmatrix} \textbf{S} \ \text{conv} 1 \times 1, 512 \\ \textbf{S} \ \text{conv} 1 \times 1, 512 \\ \textbf{S} \ \text{conv} 1 \times 1, 512 \\ \end{bmatrix}$ $\times$ 1\\
\hline
$28 \times 28$ & \ \ \! \! \! $\begin{bmatrix} \text{conv} 1 \times 1, \ 256 \\ \text{conv} 3 \times3, \ 256 \\ \text{conv} 1 \times 1, 1024 \\ \end{bmatrix}$ $\times$ 23 & \ \ \: $\begin{bmatrix} \phantom{\textbf{S}} \ \text{conv} 1 \times 1, \ 256 \\ \textbf{S} \ \text{conv} 1 \times 1, \ 256 \\ \phantom{\textbf{S}} \ \text{conv} 1 \times 1, \ 1024 \\ \end{bmatrix}$ $\times$ 23 & \ \: $\begin{bmatrix} \textbf{S} \ \text{conv} 1 \times 1, \ 256 \\ \textbf{S} \ \text{conv} 1 \times 1, \ 256 \\ \textbf{S} \ \text{conv} 1 \times 1, 1024 \\ \end{bmatrix}$ $\times$ 23 & \: $\begin{bmatrix} \textbf{S} \ \text{conv} 1 \times 1, 1024 \\ \textbf{S} \ \text{conv} 1 \times 1, 1024 \\ \textbf{S} \ \text{conv} 1 \times 1, 1024 \\ \end{bmatrix}$ $\times$ 8\\
\hline
$14 \times 14$ & \ \! \! \! $\begin{bmatrix} \text{conv} 1 \times 1, \ 512 \\ \text{conv} 3 \times3,\ 512 \\ \text{conv} 1 \times 1 , 2048 \\ \end{bmatrix}$ $\times$ \ 3 & \! \! $\begin{bmatrix} \phantom{\textbf{S}} \ \text{conv} 1 \times 1, \ 512 \\ \textbf{S} \ \text{conv} 1 \times 1, \ 512 \\ \phantom{\textbf{S}} \ \text{conv} 1 \times 1, 2048 \\ \end{bmatrix}$ $\times$ 3 & \ \: $\begin{bmatrix} \textbf{S} \ \text{conv} 1 \times 1 , \ 512 \\ \textbf{S} \ \text{conv} 1 \times 1, \ 512 \\ \textbf{S} \ \text{conv} 1 \times 1, \ 2048 \\ \end{bmatrix}$ $\times$ \ 3 & \; \! $\begin{bmatrix} \textbf{S} \ \text{conv} 1 \times 1, \ 2048 \\ \textbf{S} \ \text{conv} 1 \times 1, \ 2048 \\ \textbf{S} \ \text{conv} 1 \times 1, \ 2048 \\ \end{bmatrix}$ $\times$ 1\\
\hline
$7 \times 7$ & \multicolumn{4}{c|}{avg. pool $7 \times 7$, fc 1000, soft-max} \\
\hline
\end{tabular}
}
\caption{Overview of the architectures used in this work. Repeating residual blocks are indicated in square brackets, with the number of times the block is repeated to the right. A bold \textbf{S} indicates shift placement. For each convolution in a residual block, the number after the comma indicates the number of output channels from an operation.}%\vspace{-10pt}}
\label{tab:dimensions}
\end{table*}

%\vspace{-2pt}

Noting the results of \cite{Primitives}, we are also interested if the origin element, $a = b = 0$, is strictly necessary. In residual networks, information about the origin element can be carried by the residual connection itself. It might then be natural to not also include origin element information in shift operations used in residual networks. We first define the origin or 'no-shift' neighbourhood as:

%\vspace{-6pt}

\begin{equation}
N_{O}(\mathbf{x})= \{\mathbf{y} : \mathbf{y} = (x_{1},x_{2})\}.
\end{equation}

%\vspace{+2pt}

And then define two further shift neighbourhoods without the origin as:

%\vspace{-13pt}
\begin{eqnarray}
    N_{8C-O}(\mathbf{x})= N_{8C}(\mathbf{x}) \setminus N_{O}(\mathbf{x}),\\
    N_{4C-O}(\mathbf{x})= N_{4C}(\mathbf{x}) \setminus N_{O}(\mathbf{x}).
\end{eqnarray}
%\vspace{-13pt}

As a form of sanity check, we ask what happens if we use no shifts at all. This creates a baseline for the benefit of using shift operations when compared to a network of otherwise identical configuration. This case uses the 'no-shift' neighbourhood. Hence in total we investigate five neighbourhood variants. We apply these operations to the original ResNet \cite{ResNet}. We replace the $3 \times 3$ convolution within each residual block with a shift operation immediately followed by a point-wise $1 \times 1$ convolution. In all experiments (CIFAR100 and ImageNet) we use ResNet's 'bottleneck' residual block design (Table \ref{tab:dimensions}). The proposed changes are shown diagramatically in Fig. \ref{fig:Fig1}. We (initially) do not alter the channel structure of a block. We do this to ensure that each shift design is identical in terms of FLOPs and parameter count and can be more simply compared to ResNet.

Finally, we note a change to the downsampling method for shift-based networks. ResNet uses stride-2 spatial ($3 \times 3$) convolutions to downsample within a residual block. All input activations to this spatial convolution then contribute to its output. For shift / point-wise convolution based residual blocks, this is no longer the case: most of an input activation map's information is lost following a (shifted) stride-2 point-wise convolution. We instead use a $2\times2$ average pooling to downsample, similar to \cite{DenseNet}. We perform this pooling immediately prior to the shift / point-wise convolution, matching the downsample location to ResNet.

%\vspace{-6pt}

\subsection{Multi-stage shifting residual blocks}

In Fig. \ref{fig:Fig1} b we simply replace the spatial convolution inside each residual block with a shift operations followed by $1 \times 1$ convolution. We now add further shift operations before the down-sampling and up-sampling $1 \times 1$ point-wise convolutions (Fig. \ref{fig:Fig1} c), these convolutions previously intended to be used only for dimensionality reduction and expansion \cite{ResNet}. By adding shifts, we can add spatial information to these down- and up-sampling convolutions and expand the theoretical receptive field of each block significantly, as shown in Fig. \ref{fig:Fig1} c. The same maximum receptive field extent of three blocks is then accomplished in one block, as information from a wider area is incorporated in each block's network optimization.

Inspired by \cite{ResidualInResidual} we also add an inner residual connection, which is across the middle convolution of the residual block. Now that spatial information is also carried by the first and last convolutions of the residual block, such an inner residual connection will no-longer carry only redundant information with respect to the outer residual connection.

\subsection{Flattening the residual bottleneck}

We now look at the network channel structure in this multiple shift setting. The purpose of down- and then up-sampling within bottlenecks is to reduce the dimensionality of the spatial convolution in each residual block \cite{ResNet}. While this process reduces the amount of information processed by the spatial convolution, it also reduces computational expense. By using shifts and $1 \times 1$ convolutions, we have removed spatial convolutions from network. Shift based networks then do not have the same computational need to perform dimensionality reduction. As such, we flatten ResNet's channel structure by widening the channel count in the middle of each block to be the same as the residual (see Table \ref{tab:dimensions}). This change is motivated by the improved performance of an increased channel width in other contexts, such as in Wide ResNet\cite{WideResNet} and ResNeXT \cite{ResNeXT}. Even without spatial convolutions, this change increases the parameter count and FLOPs of the network. As the receptive field extent has also increased due to the multi-shift architecture, we reduce the length of the network to limit these costs to roughly the same as the original ResNet. Table \ref{tab:dimensions} gives an overview of the architectures of this work. 
 
%%\vspace{-8pt}

\section{Experiments and Discussion}

\textbf{Datasets:} We focus on two well-known image recognition datasets: CIFAR-100 and ImageNet. CIFAR-100 contains 50,000 training examples and 10,000 test examples for 100 classes. All images are of size $32 \times 32$. For ImageNet, we use the 1,000 classes and 1.3M images train / 50K images test split as outlined by the Large-Scale Visual Recognition Challenge (ILSVRC) \cite{ImageNet}. All images are resized to a resolution of $224 \times 224$.

\textbf{Models and training:} The initial ResNet model we take from \cite{ResNet}. For all datasets, we employ ResNet101 as a baseline, using the 'bottleneck' residual block. The same channel configuration as ResNet101 is used for all shift implementations; the computational cost is then identical across all shift based networks in the first experiments (Table \ref{tab:single}).

To train ImageNet we use an initial learning rate of 0.1 and reduce it by a factor of 10 every 30 epochs for 100 epochs in total. We use a momentum of 0.9 and a batch size of 128. In training we use a random-resized crop and a single central crop for testing following \cite{ResNet}. We test one weight decay of $4\times10^{-5}$ in the first set of ImageNet experiments (Table \ref{tab:single}), and this value and an additional weight decay value of $1\times10^{-4}$ in the second set of experiments (Table \ref{tab:multi}). Results for ImageNet are from training on 4 NVIDIA 1080Ti GPUs. Code and trained models are available at https://github.com/andrewgrahambrown/4CShiftResNet.

%\vspace{+2pt}

For training CIFAR-100, we use the same initial learning rate of 0.1 and reduce it by a factor of 10 every 100 epochs for 300 epochs. Training on a single NVIDIA TITANX GPU, we use a higher weight decay of $5\times10^{-4}$, and a batch size also of 128.

%%\vspace{-6pt}

\begin{table}[t]
\centering
\resizebox{1.0\columnwidth}{!}{
    \begin{tabular}{lccc|ccc}
    \toprule
     & \multicolumn{3}{c}{\textbf{CIFAR-100}} & \multicolumn{3}{c}{\textbf{ImageNet}}\\
     & \#params & FLOPs &  acc. & \#params & FLOPs & \makecell{acc.}\\
    \midrule
    ResNet101 \cite{ResNet} & 1078K & 154M & 74.9 & 44.6M & 7.80G & 77.6\\
    ResNet50 \cite{ResNet} & 540K & 78M & 72.3  & 25.6M & 4.09G & 75.9\\
    8-connected shift & 605K & 85M & 74.3  & 25.6M & 4.41G & 77.3\\
    4-connected shift & 605K & 85M & 73.8  & 25.6M & 4.41G & 77.3\\
    8-connected shift (nO) & 605K & 85M & 74.2 & 25.6M & 4.41G & 77.0\\
    4-connected shift (nO)& 605K & 85M & 73.5  & 25.6M & 4.41G & 77.0\\
    No shift  & 605K & 85M & 58.4  & 25.6M & 4.41G & 61.2\\
    \bottomrule
    \end{tabular}
}
\caption{Results for directly replacing spatial convolutions in ResNet101 with shifts of various neighbourhoods, compared to baselines of ResNet101 and ResNet50. nO denotes no origin. We find that 4-connected neighbourhoods are sufficient shifting in residual networks.}%\vspace{-7pt}}
\label{tab:single}
\end{table}

\subsection{Comparing shift operations}

In table \ref{tab:single} we show how the direct replacement of spatial convolutions in ResNet with different shift types affects accuracy. We first look at CIFAR100 results. When compared to the ResNet101 baseline, all investigated shifts decrease computational cost by nearly half and suffer an accuracy penalty. This penalty is however smaller than that of using a shorter ResNet of comparable computational cost, such as ResNet50. The accuracy drop is also slightly larger for those shifts not including the origin than those shifts that do include the origin. This implies that, even though the residual connection carries information about the origin, it is still necessary to also include this information within shift operations. On CIFAR100, when using only one shift within a residual block, 8-connected shifts tend to outperform 4-connected shifts. 

\begin{table*}[t]
\centering
\resizebox{0.85\columnwidth}{!}{
\begin{tabular}{lccc|cccc}
\toprule
 & \multicolumn{3}{c}{\textbf{CIFAR-100}}
 & \multicolumn{4}{c}{\textbf{ImageNet}}\\
 & \#params & FLOPs & accuracy & \#params & FLOPs & \multicolumn{2}{c}{accuracy}\\
%\makecell{accuracy wd=4e-5} & \makecell{accuracy wd=1e-4}\\
 & & & & & & wd: $4 \times 10^{-5}$ & wd: $1 \times 10^{-4}$\\
\midrule
\rowcolor{Gray}
\emph{Baselines} & & & & & & & \\
ResNet101 \cite{ResNet} & 1078K & 154M & 74.9 & 44.6M & 7.80G &  77.6 & 77.4 \\
\midrule
\rowcolor{Gray}
\emph{Multi-shifting} & & & & & & & \\

8-connected  & 605K & 85M  & 74.3 & 25.6M & 4.41G & 76.8 & 77.2\\
4-connected  & 605K & 85M  & 75.1 & 25.6M & 4.41G & 77.3 & 77.6\\
\midrule
\rowcolor{Gray}
\emph{Flattened architecture} & & & & & & & \\
8-connected & 1068K & 162M & 76.9 & 40.8M & 7.72G & 77.2 & 77.8\\
4-connected & 1068K & 162M & 77.5 & 40.8M & 7.72G & 77.8 & 78.4\\
\bottomrule
\end{tabular}%
}
\caption{Results for networks with additional shifts placed before down- and up-sampling convolutions, compared to the baseline ResNet101. For both datasets we find that, when using multiple shifts, 4-connected shifts are preferred over 8-connected shifts. The accuracies of multiple 4-connected shift networks are competitive with the baseline at a reduced computational cost. Using multiple shifts in a flattened residual block channel structure results in an improved performance over standard ResNets at a similar computational cost. In this flattened architecture, we again find 4-connected shifts are preferred over 8-connected shifts.} %\vspace{-5pt}}
\label{tab:multi}
\end{table*}

Similar results on are seen on ImageNet as on CIFAR-100: using shifts reduces computational cost, but an accuracy penalty is suffered. Noting that the absolute accuracies on both CIFAR-100 and ImageNet are similar, this penalty is smaller for ImageNet, between -0.3\% and -0.6\%, than for CIFAR-100, between -0.6\% and -1.4\%. We again find that shifts with an origin component outperform those without an origin component. One important difference between CIFAR-100 and ImageNet results is that 4C shifts show equal performance to 8C shifts. This result is unexpected, as the theoretical size of the receptive field is restricted for 4C shifts when compared to 8C shifts (Fig. \ref{fig:Fig1}).

Lastly, we note that for both CIFAR-100 and ImageNet, we find that having no shift at all drops accuracy significantly, but only to 58.4\% and 61.2\% for each dataset respectively. That network accuracy remains this high is surprising: these networks have only a single spatial convolution in their first layer. All other convolutions are point-wise $1 \times 1$ and cannot include spatial information (Table \ref{tab:dimensions}) - yet accuracy is still high enough to beat AlexNet \cite{AlexNet}. Such no-shift networks are similar in structure to BagNets \cite{BagNet} - networks which have a highly restricted set of spatial convolutions. The most important distinction is that our no-origin networks do not include \textit{any} spatial convolutions beyond the first convolution. Comparatively, BagNets still include one additional spatial convolution in each of ResNet's four layers. Our networks then have a greater spatial extent restriction than BagNets. Our results then suggest that perceptual tasks such as ImageNet can be solved by even smaller spatial feature extents than previous shown in \cite{BagNet}.

In the next section, we examine the effects of placing shifts at additional positions in the network. We do this for both the best performing shifts on ImageNet from this section, the 8C and 4C shifts including an origin component.

\subsection{Multi-stage shifting}

In Table \ref{tab:multi} we show networks with multiple shift operations in each residual block and compare them to a baseline ResNet101, again on both CIFAR-100 and ImageNet. 

For CIFAR-100, we find that the multi-4C shift networks improves against single-4C shift networks (+1.2\%), but multi-8C shift networks show no improvement over single-8C shift networks. The accuracy for multi-4C shift networks is slightly above the baseline (+0.2\%), while reducing computational costs by 45\%. The final architecture studied flattens the channel structure of bottlenecks and has a reduced network length, keeping computational costs approximately the same as the baseline ResNet101. In this architecture, 4C shifts are again found to outperform 8C shifts. Both shift types outperform the baseline in this architecture, with 4C shifts giving the greatest accuracy improvement (+2.6\%).

For ImageNet models we compare two weight decay settings: $4\times10^{-5}$, suggested in \cite{SqueezeExcite} for use with ResNet architectures, and $1\times10^{-4}$, used in the original ResNet experiments \cite{ResNet}. We find that multi-shift networks are particularly sensitive to weight decay within this range. All multi-shift networks benefit from using the same weight decay as originally suggested for ResNet, though ResNet itself does not. While not shown in Table \ref{tab:single}, a higher weight decay degrades performance for single shifts. In both weight decay settings, we find that multiple 4C shifts outperform multiple 8C shifts. This is despite the reduced theoretical receptive field size of 4C shifts when compared to 8C shifts. Comparing optimal weight decay settings for each network, adding multiple shift modules improves 4C shift results (+0.3\%), but does not change 8C results. The multi-4C shift architecture provides the same accuracy as the original ResNet101, yet with a 43\% reduction in computational costs.

%Table \ref{tab:multi} also shows results from a network with multiple 4C shifts and a flattened channel structure. This network significantly improves accuracy (+0.8\%) with respect to a baseline ResNet101 for approximately the same computation. This improved accuracy is in spite of these shift-based networks being considerably less deep (35 layers) than the baseline (101 layers), see Table \ref{tab:dimensions}. This choice of depth was made to keep the FLOPs and parameter count approximately the same as ResNet, and does not appear to have restricted accuracy.  

Table \ref{tab:multi} also shows ImageNet results from networks with a flattened channel structure and equipped with either multiple 4C shifts or multiple 8C shifts. We also find that in this architecture, multiple 4C shift out-perform multiple 8C shifts. In this architecture, using either 8C or 4C shifts results in an improved accuracy against the baseline ResNet101 while keeping computational cost approximately the same, with use of 4C shifts yielding the largest improvement (+0.8\%). This improved accuracy is in spite of these shift-based networks being considerably less deep (35 layers) than the baseline (101 layers), see Table \ref{tab:dimensions}. This choice of depth was made to keep the FLOPs and parameter count approximately the same as ResNet, and does not appear to have restricted accuracy.  

\begin{figure}
\begin{center}
%\floatbox[{\capbeside\thisfloatsetup{capbesideposition={right,top},capbesidewidth=0.4\textwidth}}]{figure}[\FBwidth]
%{\includegraphics[width=\linewidth]{images/FigComparisonV2.png}}
%{
\includegraphics[width=\linewidth]{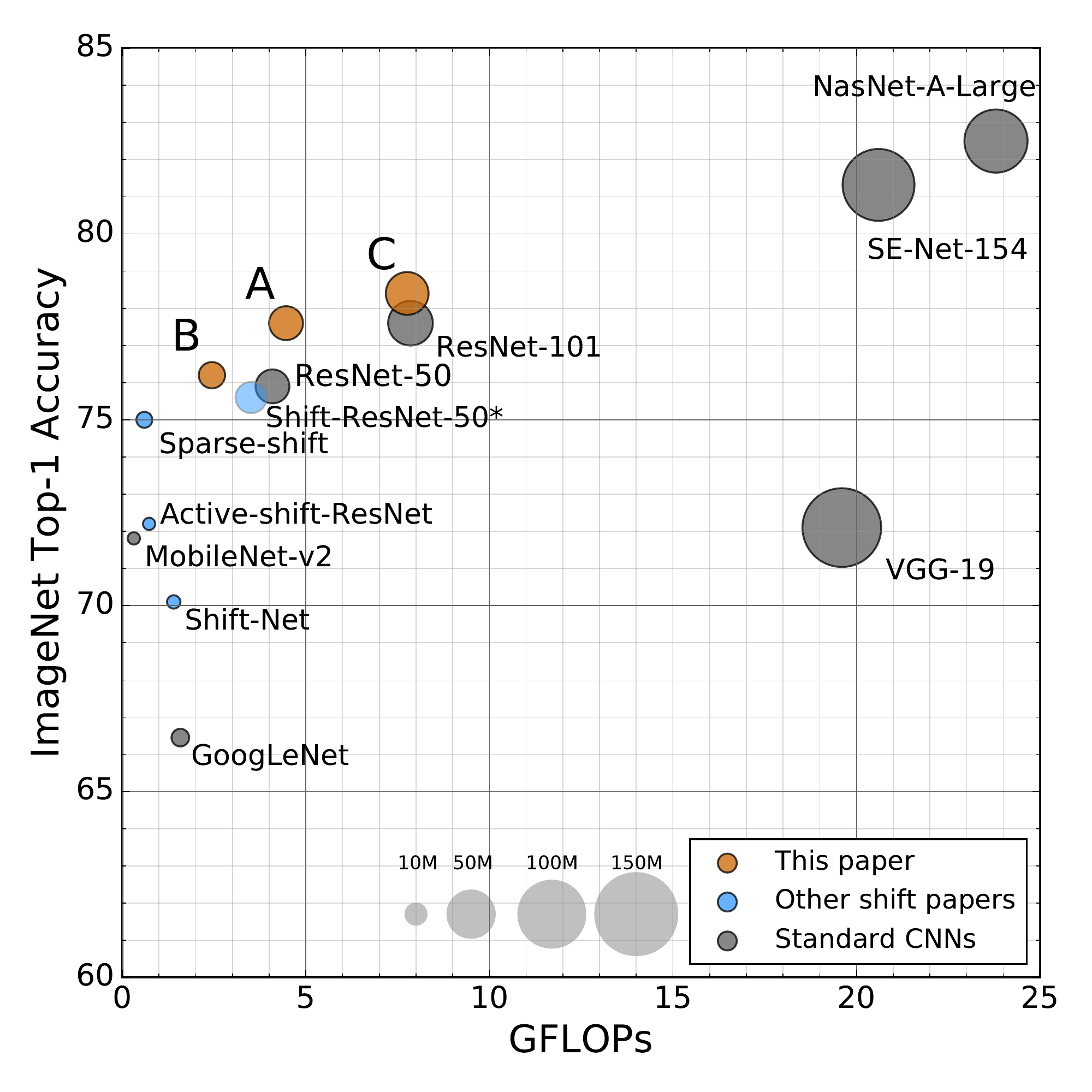}
\caption{ImageNet top-1 accuracies as they relate to FLOPs, with the parameter count indicated by circle size. See text for data sources. Our approach (orange circles) demonstrates shifts can improve FLOPs/parameters or accuracy against ResNet. 'A' and 'B' denote 4C-MS-ResNet101 and 4C-MS-ResNet50. Both are models using multiple shifts with the original ResNet channel structure. 'C' denotes 4C-MSF-ResNet35 which uses the flattened channel structure. All variants use 4-connected shifts. In terms of accuracy, our networks outperform the popular CNN architectures VGG-19 \cite{VGG} and MobileNetv2 \cite{MobileNetV2} and all other shift-based networks \cite{Active,Zero}. The FLOPs for Shift-ResNet-50 \cite{Zero} are not available and have been estimated from the parameter count.}
\label{fig:Fig3}
%}
\end{center}
%\vspace{-30pt}
\end{figure}

In our final figure, Fig.~\ref{fig:Fig3}, we comparatively evaluate the top1-accuracy on ImageNet of different networks as a function of both the number of FLOPs and the number of network parameters. The figure shows how our multi-4C shift residual network design significantly improves in computational cost against one of the most popular modern network designs, ResNet \cite{ResNet}, while maintaining accuracy. On the other hand, our flattened multi-4C shift architecture has a similar numbers of FLOPs and parameters as ResNet and improves accuracy.

We draw a comparison to shift papers with ImageNet architectures \cite{Active,Zero,Sparse}. We highlight that while the networks shown in these works are computationally efficient, their accuracies are comparatively lower. This is as these works principally focused on improving compact, low parameter / FLOP networks; their results can thus be seen towards the left of the figure. The exception is ShiftResNet50 shown in \cite{Zero}. The exact architecture and FLOPs of this network were not reported; here we have estimated the network's FLOPs from the number of reported parameters and show the results in Fig.~\ref{fig:Fig3}.

%\newpage

 We also compare to well established standard CNN architectures \cite{GoogleNet,VGG,SqueezeExcite,NasNet,MobileNetV2}. For the accuracies, FLOPs and parameters of standard CNNs, we use the benchmark analysis of Bianco et al. \cite{Benchmark}. Compared to other standard CNN architecures, the accuracy of our networks are superior to MobileNetv2 \cite{MobileNetV2}, GoogleNet \cite{GoogleNet} and VGG \cite{VGG}. The current best performers on ImageNet, SENet-154 \cite{SqueezeExcite} and NasNet-A-Large \cite{NasNet}, have a higher accuracy than our networks, but come with a much larger FLOP and parameter demand. We envision that these networks can similarly benefit from using 4-connected shifts in their architecture, reducing their FLOP requirement yet maintaining accuracy.

%\vspace{-30pt}

\section{Conclusions}

This work investigates shifts in deep residual networks and how best to apply them in the high accuracy, large image classification setting. We examine shifts based on both the 8-connected and 4-connected neighbourhoods. We find that, when used solely within residual blocks, both neighbourhoods offer similar performance. When used multiple times, the shift neighbourhood should be restricted to the 4-connected neighbours. As such, we posit that only shifting to the 4 nearest neighbours is sufficient in deep residual networks. We have outlined two high-accuracy networks using 4-connected shifts: the first reduces computational cost against ResNet101 by 43\% without compromising on accuracy; the second improves on ResNet101's accuracy, while keeping computational costs roughly equal. These results show that shifts can be successfully applied in the high-accuracy deep learning setting, offering large improvements in computational cost or accuracy. Code and trained models are available at https://github.com/andrewgrahambrown/4CShiftResNet.

{\small
\bibliographystyle{ieee_fullname}
\bibliography{egbib}

\begin{thebibliography}{10}\itemsep=-1pt

\bibitem{Benchmark}
Simone Bianco, Remi Cadene, Luigi Celona, and Paolo Napoletano.
\newblock Benchmark analysis of representative deep neural network
  architectures.
\newblock {\em IEEE Access}, 6:64270--64277, 2018.

\bibitem{BagNet}
Wieland Brendel and Matthias Bethge.
\newblock Approximating cnns with bag-of-local-features models works
  surprisingly well on imagenet.
\newblock In {\em ICLR}, 2019.

\bibitem{TensorFactorisation}
Nadav Cohen, Or Sharir, and Amnon Shashua.
\newblock On the expressive power of deep learning: A tensor analysis.
\newblock In {\em COLT}, 2016.

\bibitem{TensorFactorisation2}
Emily~L. Denton, Wojciech Zaremba, Joan Bruna, Yann LeCun, and Rob Fergus.
\newblock Exploiting linear structure within convolutional networks for
  efficient evaluation.
\newblock In {\em NeurIPS}. 2014.

\bibitem{8vs4_3}
Luigi Di~Stefano and Andrea Bulgarelli.
\newblock A simple and efficient connected components labeling algorithm.
\newblock In {\em ICIAP}, pages 322--327, 1999.

\bibitem{Cellular}
M.J.B. Duff, D.M. Watson, T.J. Fountain, and G.K. Shaw.
\newblock A cellular logic array for image processing.
\newblock {\em Pattern Recognition}, 5(3):229 -- 247, 1973.

\bibitem{TensorFlow}
Mart\'in~Abadi et al.
\newblock Tensorflow: Large-scale machine learning on heterogeneous distributed
  systems, 2016, arXiv:1603.04467.

\bibitem{NetworkPruning2}
Yiwen Guo, Anbang Yao, and Yurong Chen.
\newblock Dynamic network surgery for efficient dnns.
\newblock In {\em NeurIPS}, 2016.

\bibitem{NetworkPruning}
Song Han, Jeff Pool, John Tran, and William~J. Dally.
\newblock Learning both weights and connections for efficient neural networks.
\newblock In {\em NeurIPS}, 2015.

\bibitem{ResNet}
Kaiming He, Xiangyu Zhang, Shaoqing Ren, and Jian Sun.
\newblock Deep residual learning for image recognition.
\newblock In {\em CVPR}, 2016.

\bibitem{Primitives}
Yihui He, Xianggen Liu, Huasong Zhong, and Yuchun Ma.
\newblock Addressnet: Shift-based primitives for efficient convolutional neural
  networks.
\newblock In {\em WACV}, 2019.

\bibitem{StudentTeacher}
Geoffrey Hinton, Oriol Vinyals, and Jeffrey Dean.
\newblock Distilling the knowledge in a neural network.
\newblock In {\em NeurIPS workshop}, 2015.

\bibitem{SqueezeExcite}
Jie Hu, Li Shen, and Gang Sun.
\newblock Squeeze-and-excitation networks.
\newblock In {\em CVPR}, 2018.

\bibitem{DenseNet}
Gao Huang, Zhuang Liu, Laurens Van Der~Maaten, and Kilian~Q Weinberger.
\newblock Densely connected convolutional networks.
\newblock In {\em CVPR}, 2017.

\bibitem{Squeeze}
Forrest~N. Iandola, Matthew~W. Moskewicz, Khalid Ashraf, Song Han, William~J.
  Dally, and Kurt Keutzer.
\newblock Squeezenet: Alexnet-level accuracy with 50x fewer parameters and less
  than 1mb model size, 2016, arXiv:1602.07360.

\bibitem{Active}
Yunho Jeon and Junmo Kim.
\newblock Constructing fast network through deconstruction of convolution.
\newblock In {\em NeurIPS}, 2018.

\bibitem{CIFAR}
Alex Krizhevsky.
\newblock Learning multiple layers of features from tiny images.
\newblock Technical report, 2009.

\bibitem{AlexNet}
Alex Krizhevsky, Ilya Sutskever, and Geoffrey~E Hinton.
\newblock Imagenet classification with deep convolutional neural networks.
\newblock In {\em NeurIPS}, 2012.

\bibitem{Systolic}
Hsiang-Tsung Kung, Bradley McDanel, and Sai~Quian Zhang.
\newblock Mapping systolic arrays onto 3d circuit structures: Accelerating
  convolutional neural network inference.
\newblock In {\em SiPS}, 2018.

\bibitem{MotionFeatureNetwork}
Myunggi Lee, Seungeui Lee, Sungjoon Son, Gyutae Park, and Nojun Kwak.
\newblock Motion feature network: Fixed motion filter for action recognition.
\newblock In {\em ECCV}, 2018.

\bibitem{Pytorch}
Adam Paszke, Sam Gross, Soumith Chintala, Gregory Chanan, Edward Yang, Zachary
  DeVito, Zeming Lin, Alban Desmaison, Luca Antiga, and Adam Lerer.
\newblock Automatic differentiation in pytorch.
\newblock In {\em NeurIPS workshop}, 2017.

\bibitem{CVHandbook}
Gerhard Ritter and Joseph Wilson.
\newblock {\em Handbook of Computer Vision Algorithms in Image Algebra}.
\newblock CRC Press, Inc., 1996.

\bibitem{8vs4_1}
Jos Roerdink and A Meijster.
\newblock The watershed transform: Definitions, algorithms and parallelization
  strategies.
\newblock {\em Fundamental Informatica}, 41, 10 2003.

\bibitem{StudentTeacher2}
Adriana Romero, Nicolas Ballas, Samira~Ebrahimi Kahou, Antoine Chassang, Carlo
  Gatta, and Y Bengio.
\newblock Fitnets: Hints for thin deep nets, 2014,arXiv:1412.6550.

\bibitem{ImageNet}
Olga Russakovsky, Jia Deng, Hao Su, Jonathan Krause, Sanjeev Satheesh, Sean Ma,
  Zhiheng Huang, Andrej Karpathy, Aditya Khosla, Michael Bernstein, et~al.
\newblock Imagenet large scale visual recognition challenge.
\newblock {\em IJCV}, 115(3):211--252, 2015.

\bibitem{MobileNetV2}
Mark Sandler, Andrew Howard, Menglong Zhu, Andrey Zhmoginov, and Liang-Chieh
  Chen.
\newblock Mobilenetv2: Inverted residuals and linear bottlenecks.
\newblock In {\em CVPR}, 2018.

\bibitem{Serra}
Jean Serra.
\newblock {\em Image Analysis and Mathematical Morphology}.
\newblock Academic Press, Inc., 1983.

\bibitem{VGG}
Karen Simonyan and Andrew Zisserman.
\newblock Very deep convolutional networks for large-scale image recognition.
\newblock In {\em ICLR}, 2015.

\bibitem{GoogleNet}
Christian Szegedy, Wei Liu, Yangqing Jia, Pierre Sermanet, Scott Reed, Dragomir
  Anguelov, Dumitru Erhan, Vincent Vanhoucke, and Andrew Rabinovich.
\newblock Going deeper with convolutions.
\newblock In {\em CVPR}, 2015.

\bibitem{8vs4_2}
Lucas~J. van Vliet and Ben~J.H. Verwer.
\newblock A contour processing method for fast binary neighbourhood operations.
\newblock {\em Pattern Recognition Letters}, 7(1):27 -- 36, 1988.

\bibitem{Sparse}
Chen Weijie, Di Xie, Yuan Zhang, and Shiliang Pu.
\newblock All you need is a few shifts: Designing efficient convolutional
  neural networks for image classification.
\newblock In {\em CVPR}, 2019.

\bibitem{Zero}
Bichen Wu, Alvin Wan, Xiangyu Yue, Peter Jin, Sicheng Zhao, Noah Golmant, Amir
  Gholaminejad, Joseph Gonzalez, and Kurt Keutzer.
\newblock Shift: A zero flop, zero parameter alternative to spatial
  convolutions.
\newblock In {\em CVPR}, 2018.

\bibitem{ResNeXT}
Saining Xie, Ross Girshick, Piotr Dollar, Zhuowen Tu, and Kaiming He.
\newblock Aggregated residual transformations for deep neural networks.
\newblock In {\em CVPR}, 2017.

\bibitem{Synetgy}
Yifan et~al. Yang.
\newblock Synetgy: Algorithm-hardware co-design for convnet accelerators on
  embedded fpgas.
\newblock In {\em International Symposium on Field-Programmable Gate Arrays},
  2018.

\bibitem{WideResNet}
Sergey Zagoruyko and Nikos Komodakis.
\newblock Wide residual networks.
\newblock In {\em BMVC}, 2016.

\bibitem{ResidualInResidual}
Ke Zhang, Miao Sun, Tony~X. Han, Xingfang Yuan, Liru Guo, and Tao Liu.
\newblock Residual networks of residual networks: Multilevel residual networks.
\newblock {\em TCSVT}, 28(6), 2018.

\bibitem{NasNet}
Barret Zoph, Vijay Vasudevan, Jonathon Shlens, and Quoc~V. Le.
\newblock Learning transferable architectures for scalable image recognition.
\newblock In {\em CVPR}, 2018.

\end{thebibliography}
}

\end{document}